\documentclass[11pt,a4paper]{article}
\usepackage[hyperref]{naaclhlt2019}
\usepackage{times}
\usepackage{latexsym}

\usepackage{url}

\aclfinalcopy 

\newcommand\tablewidth{0.92}

\usepackage{amsfonts}
\usepackage{amsmath}

\usepackage{graphicx}

\usepackage{algorithm}
\usepackage{algorithmic}

\usepackage{tabularx}

\usepackage[inline, shortlabels]{enumitem}

\usepackage{multirow}

\usepackage{placeins}
\usepackage{blindtext}

\usepackage{float}

\usepackage{tikz}
\usetikzlibrary{shapes,snakes}
\usetikzlibrary{arrows, arrows.meta, calc, snakes, positioning, quotes, shapes}

\usepackage{graphicx}
\usepackage[labelformat=simple]{subcaption}

\usepackage{fixltx2e}

\title{DAL: Dual Adversarial Learning for Dialogue Generation}

\author{ Shaobo Cui~\textsuperscript{1}, Rongzhong Lian~\textsuperscript{2}, Di Jiang~\textsuperscript{2}, Yuanfeng Song~\textsuperscript{2}, Siqi Bao~\textsuperscript{2}, Yong Jiang~\textsuperscript{1}\\
  \textsuperscript{1}Tsinghua University, China \\
   \textsuperscript{2}Baidu Inc., China \\
  {\tt cuishaobo16@mails.tsinghua.edu.cn } \\ 
  {\tt \{lianrongzhong,jiangdi,songyuanfeng,baosiqi\}@baidu.com} \\
  {\tt jiangy@sz.tsinghua.edu.cn}
  }

\date{}

\begin{document}
\maketitle
\begin{abstract}
In open-domain dialogue systems, generative approaches have attracted much attention for response\footnote{We use \textit{query} and \textit{response} to denote the first and second utterances in a single-turn dialogue. } generation. However, existing methods are heavily plagued by generating safe responses and unnatural responses. To alleviate these two problems, we propose a novel framework named Dual Adversarial Learning~(DAL) for high-quality response generation. DAL is the \textbf{first} work to innovatively utilizes the duality between query generation and response generation to avoid safe responses and increase the diversity of the generated responses. Additionally, DAL uses adversarial learning to mimic human judges and guides the system to generate natural responses. Experimental results demonstrate that DAL effectively improves both diversity and overall quality of the generated responses. DAL outperforms the state-of-the-art methods regarding automatic metrics and human evaluations.
\end{abstract}
\section{Introduction}
In recent years, open-domain dialogue systems are gaining much attention owing to their great potential in applications such as educational robots, emotional companion, and chitchat. The existing approaches for open-domain dialogue systems can be divided into two categories: retrieval-based approaches~\cite{hu2014convolutional,ji2014information} and generative approaches~\cite{ritter2011data,shang2015neural}. The retrieval-based approaches are based on conventional information retrieval techniques and strongly rely on the underlying corpus~\cite{wang2013dataset,lu2013deep}. Since the capability of retrieval-based approaches is strongly limited by corpus, generative approaches are becoming more and more prominent in the field of open-domain dialogue research. The \emph{de facto} backbone of generative approaches is the Seq2Seq model~\cite{bahdanau2014neural} , which is essentially an encoder-decoder neural network architecture. Despite their success, Seq2Seq model and its variants~\cite{sordoni2015neural,vinyals2015neural} are heavily plagued by \textbf{safe responses}~(generic and dull responses such as ``I don't know" or ``Me too") and \textbf{unnatural responses}~(such as ``I want to go, but I don't want to go").

In this paper, we propose a novel framework named Dual Adversarial Learning~(DAL) to alleviate the aforementioned two problems. DAL consists of two generative adversarial networks~(GANs): one for query generation and the other for response generation.
The response generation model is used to transfer from the query domain $\mathcal{Q}$ to the response domain $\mathcal{R}$, while the query generation model is for transformation from $\mathcal{R}$ to $\mathcal{Q}$. Here we consider the response generation task and the query generation task as \textbf{dual} tasks. 
 The generators of these two GANs are connected through the duality constraint.  
As such, in DAL, there are two kinds of signals that jointly instruct the optimization of generators:  (1) the dual signal from the duality constraint between these two generators;  (2) the adversarial signal from the discriminators.
The dual signal is utilized to model the mutual relation between query generation and response generation. 	 
We use an instance to better illustrate this mutual relation: for a given query ``Where to have dinner?", compared with a safe response ``I don’t know", a more diverse and specific response ``The Indian cuisine around the corner is great" usually has a higher probability of being transformed back to the given query.
DAL takes full advantage of this intuition via dual learning, which avoids generating safe responses and improves the diversity of the generated responses. 
Additionally, in order to make the generated responses as natural as possible, the adversarial signal in DAL mimics human judges to alleviate unnatural responses.
We compare DAL with state-of-the-art methods through extensive experiments, and DAL demonstrates superior performance regarding automatic metrics, human evaluations, and efficiency.

There are \textbf{crucial differences} between our dual approach and Maximum Mutual Information~(MMI)~\cite{li2016diversity} though both utilize the reverse dependency to improve the diversity of the generated responses. Due to the challenging mutual information objective, the distribution $p(r \vert q)$ is same as that in vanilla Seq2Seq in MMI. More specifically, $p(r|q)$ in MMI is trained only by maximum likelihood objective at training time~({we use $p(r \vert q)$ to denote the probability distribution of predicting the response $r$ given the query $q$}). The mutual information in MMI is utilized only at inference time, and the inference process is not only time-consuming but also inaccurate in MMI. 
However, $p(r \vert q)$ in our dual approach is trained by not only the maximum likelihood objective but also the diversity objective~(duality constraint) at training time.  Since the dual approach directly incorporates the reverse dependency information at the training time, it can avoid the time-consuming inference plaguing MMI. 
Additionally, the dual approach does not need to maintain a large size optional response set for the time-consuming reranking strategy in MMI-bidi~(one variant of MMI). The dual approach shows its efficiency superiority over MMI in real-life applications, which is shown in our efficiency experiment. 
 
Our dual approach is quite different from the reinforcement learning based structure having two Seq2Seq models in \cite{zhang2018reinforcing}\footnote{Our dual approach is finished independently with this work in addition to the crucial difference. We did not notice this paper until our work is done.}. In \cite{zhang2018reinforcing}, $G_1$, which generates a response $\hat{r}$ given a query $q$, uses the conditional probability $P_{2}(q \vert \hat{r})$ calculated by $G_{2}$ as the coherence measure to guide $G_{1}$ in the reinforcement learning process.  Similarly, $G_{2}$, which generates a query $\hat{q}$ given a response $r$, uses the conditional probability $P_{1}(r \vert \hat{q})$ calculated by $G_1$ as the coherence measure to guide $G_{2}$ in the reinforcing learning process.  
However, in our work, we utilize the joint probability $p(q,r)$ to connect these two Seq2Seq models and thus avoid unstable and time-consuming reinforcement learning in the dual approach. 
Recently, we notice that \citeauthor{zhang2018generating} \shortcite{zhang2018generating} propose to use an adversarial learning method named adversarial information maximization~(AIM) to improve the informativeness and diversity of generated responses. Though AIM also uses two reverse models, it involves the calculation of $p(\hat{r} \vert q)$ and $p(\hat{q} \vert r)$. Our DAL, however, involves the calculation of $p(r \vert q)$ and $p(q \vert r)$. As for the model structure, those two reverse Seq2Seq models in AIM share the discriminator while the two reverse Seq2Seq models in DAL have their own discriminators. The reason for these difference is that the objective of our DAL is to enforce the dual constraint while AIM is to use adversarial learning method  to optimize a variational lower bound on mutual information between query and response. 
Besides, our DAL framework is strongly different from previous structures that are composed of two GANs, such as CycleGAN~\cite{zhu2017unpaired}, DiscoGAN~\cite{kim2017learning} and DualGAN~\cite{yi2017dualgan}. Those works can \text{only} be utilized on the image translation task and two generators are connected by \textit{cycle consistency}, i.e., {for each image $x$ in domain $\mathcal{X}$, the image translation cycle is supposed to bring $x$ to the original image: $x \rightarrow G_1(x) \rightarrow G_2(G_1(x)) \approx x$ }. However, \textit{cycle consistency} is difficult to be applied into the text generation task.  In our paper, we use the \textit{joint distribution} of query-response pairs rather than \textit{cycle consistency} to enforce the duality between these two dual generators.

The contributions of this paper are listed as follows:

\noindent$\bullet$ To the best of our knowledge, this is the \textbf{first work} that adopts the duality to avoid safe responses in open-domain dialogue systems. 
It sheds light on the utility of query generation in improving the performance of response generation.

\noindent$\bullet$ DAL is a novel framework that integrates dual learning and adversarial learning, which complementary and jointly contributes to generating both diverse and natural responses. 

The rest of this paper is organized as follows. The related work is firstly reviewed. The DAL framework is introduced in Section~\ref{sect:framework} and the training of DAL is described in Section~\ref{sect:training}. Experimental results are shown in Section~\ref{sect:experiment}, followed by the conclusion of this paper in Section~\ref{sect:conclusion}.
\section{Related Work} \label{sect:related_work}
\begin{figure*}
\centering
\begin{subfigure}[b]{0.90\columnwidth}
\centering
\includegraphics[width=\linewidth]{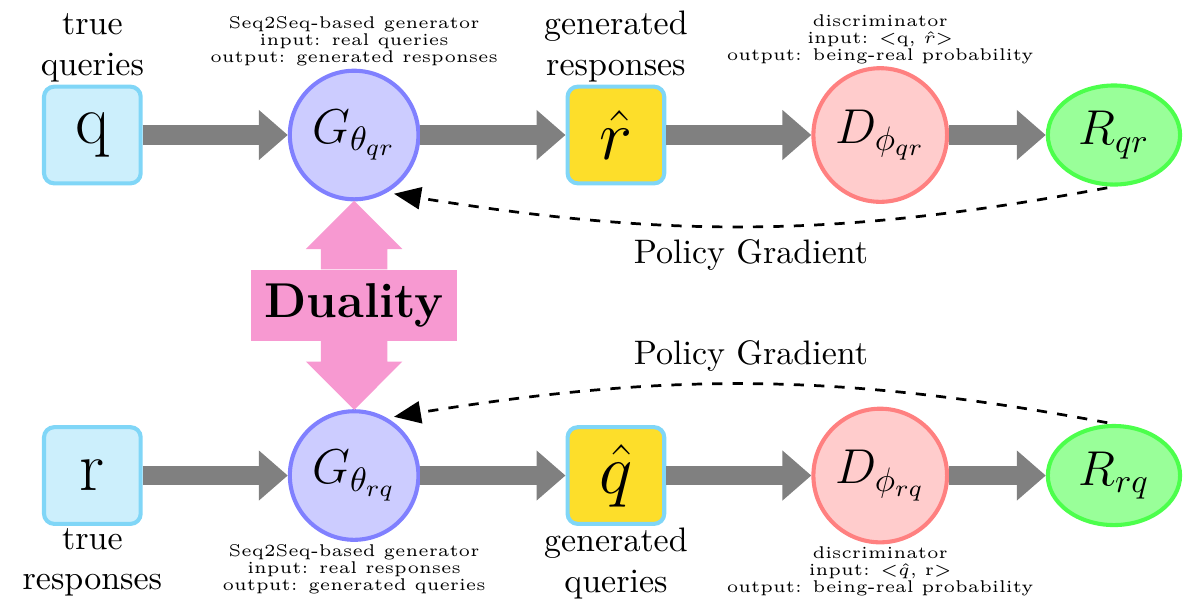}
\caption{\label{fig:double_D_model} The architecture of DAL.}
\end{subfigure}
~
\centering
\begin{subfigure}[b]{0.90\columnwidth}
	\centering
	\includegraphics[width=\linewidth]{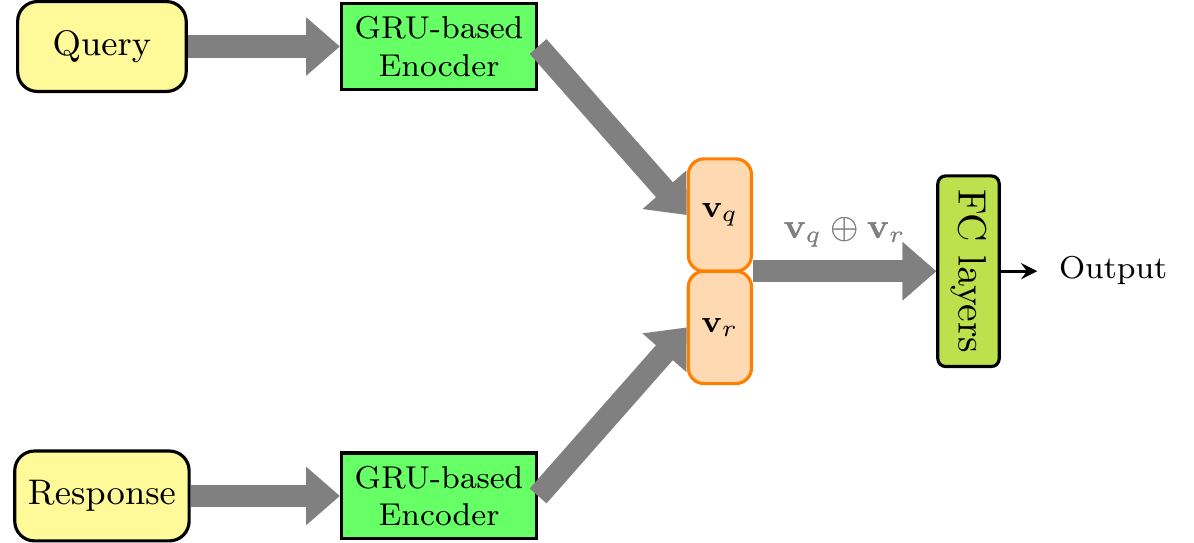}
	\caption{The architecture of the discriminator.}
	\label{fig:model:dis}
\end{subfigure}
\caption{Dual Adversarial Learning.}
\end{figure*}
\subsection{Dual Learning}
\label{sect:related_work:dual}
Many machine learning tasks have emerged in dual forms, such as dual neural machine translation~(dual-NMT)~\cite{he2016dual}, image classification and conditional image generation~\cite{van2016conditional}. 
Dual learning~\cite{he2016dual} is proposed on the assumption that the dual correlation could be used to improve both the primal task and its dual task: the primal task aims to map from input space $\mathcal{X}$ to output space $\mathcal{Y}$, whereas the dual task takes samples from space $\mathcal{Y}$ and maps to space $\mathcal{X}$. \citeauthor{tang2017question} \shortcite{tang2017question} implemented a dual framework for the question answering system. Their model regards the answer selection~(given a question and its several candidate answers, select the most satisfying answer to answer the question) and the question generation as dual tasks, which increases the performance of both.
\subsection{Adversarial Learning}
Adversarial learning~\cite{goodfellow2014generative}, or Generative Adversarial Network~(GAN), has been proven to be a promising approach for generation tasks. GAN achieves great success on the image generation task~\cite{huang2017stacked}. However, since the decoding phase in the Seq2Seq model involves sampling discrete words, GAN cannot be directly applied to the generative approach for text generation. By regarding the sequence generation as an action-taking problem in reinforcement learning, \citeauthor{li2017adversarial} \shortcite{li2017adversarial} proposed to apply GAN to dialogue generation, in which the output of the discriminator is used as the reward for the generator's optimization. \citeauthor{xu2017neural} \shortcite{xu2017neural} introduced an approximate embedding layer to solve the non-differentiable problem caused by the discrete decoding phase. 
\subsection{Work on the Safe Response Problem}
\label{sect:related_work:safe}
There is some existing work on the safe response problem. The first kind of approach is to introduce specific keywords~\cite{mou2016sequence} or topic information~\cite{xing2017topic} into the generated responses. These methods shift the difficulty from diverse response generation to keyword or topic prediction, which are also challenging tasks. 
The second kind of approach takes the reverse dependency~(the query generation task given the responses) into consideration. \citeauthor{li2016diversity} \shortcite{li2016diversity} considered the reverse dependency and proposed Maximum Mutual Information~(MMI) method, which is empirically plagued by ungrammatical responses~(MMI-antiLM) and huge decoding space~(MMI-bidi).  
\section{DAL Framework} \label{sect:framework}
In this section, we firstly give an overview of the DAL framework in Section~\ref{sect:model:overview} and then elaborate the discriminators  and dual generators in Section~\ref{sect:model:dis} and Section~\ref{sect:model:gen} separately. Finally, we discuss the reason why duality promotes diversity in Section~\ref{sect:model:dual}.
\subsection{Overview} \label{sect:model:overview}
The architecture of DAL is presented in Figure~\ref{fig:double_D_model}.
The real query and response are denoted by $q$ and $r$, whereas the generated query and response are denoted as $\hat{q}$ and $\hat{r}$.
DAL consists of two GANs~(one for query generation and the other for response generation).
Generators are denoted by $G_{\theta_{qr}}$ and $G_{\theta_{rq}}$ and the corresponding discriminators are denoted as $D_{\phi_{qr}}$ and $D_{\phi_{rq}}$.
The input of $G_{\theta_{qr}}$ is a real query $q$ and the output is the generated response $\hat{r}$.
Similarly, for $G_{\theta_{rq}}$, the input is a real response $r$ and the output is the generated query $\hat{q}$.
For $D_{\phi_{qr}}$, the input is the \textit{ficto-facto} query-response pair $\langle{q, \hat{r}}\rangle$, and the output $R_{qr}$ is estimated probability of the query-response pair being human-generated, which is estimated by $D_{\phi_{qr}}$.
Analogously, the input of $D_{\phi_{rq}}$ is the \textit{ficto-facto} pair $\langle{\hat{q}, r}\rangle$, and the output $R_{rq}$ is the estimated probability of the input pair being human-generated.
$G_{\theta_{qr}}$ and $G_{\theta_{rq}}$ are connected by the duality constraint derived from the joint probability $P(q, r)$. The adversarial signal from discriminators, $R_{qr}$, $R_{rq}$, are passed to the corresponding generators as the reward through policy gradient.
\begin{figure*}
\centering
\begin{subfigure}[b]{0.90\columnwidth}
\centering
\includegraphics[width=\linewidth]{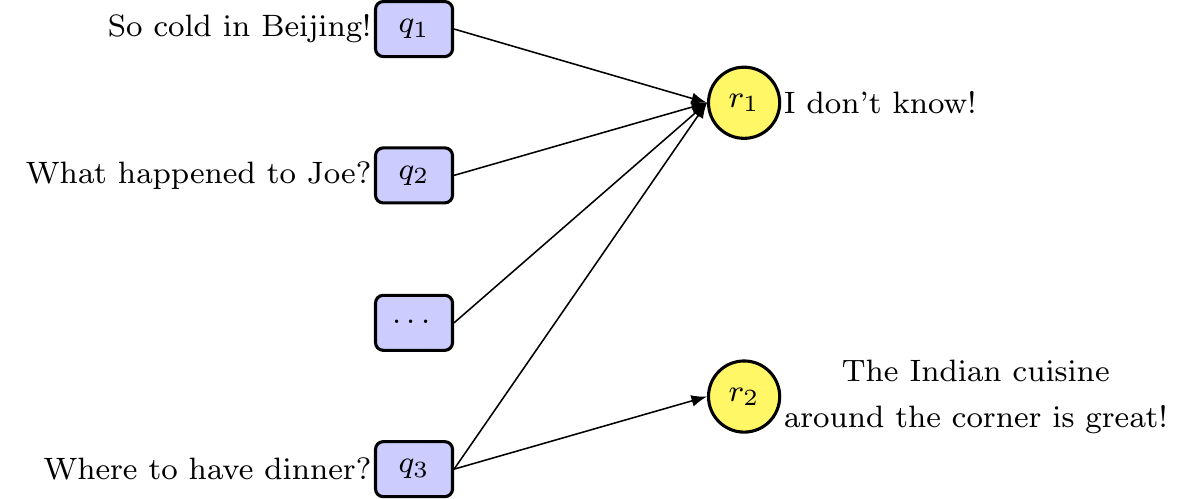}
		\caption{An example corpus.}
		\label{fig:dual:corpus}
\end{subfigure}
~
\begin{subfigure}[b]{0.90\columnwidth}
	\centering
	\includegraphics[width=\linewidth]{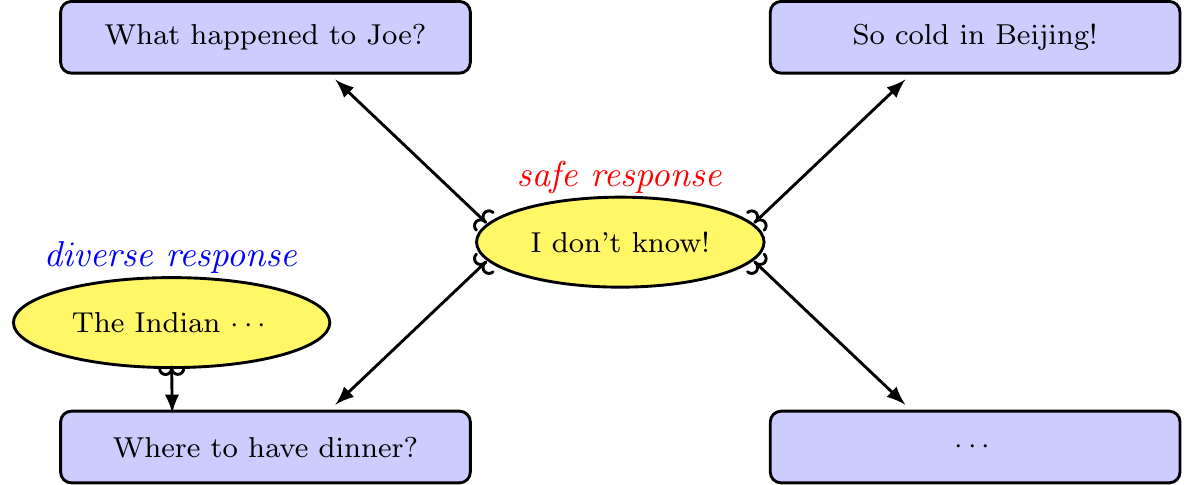}
		\caption{Queries and responses with duality constraint.}
		\label{fig:dual:graph}	
\end{subfigure}
	\caption{An example to illustrate why duality promotes diversity.}
	\label{fig:dual}
\end{figure*}
\subsection{Discriminator} \label{sect:model:dis}
The discriminator mimics a human judge and guides the generator to generate natural utterances. The architecture of the discriminator is shown in Figure~\ref{fig:model:dis}.
Gated Recurrent Unit~(GRU) based~\cite{bahdanau2014neural} neural networks are used to obtain the query embedding $\mathbf{v}_q$ and the response embedding $\mathbf{v}_r$. The concatenation vector $\mathbf{v}_q \oplus \mathbf{v}_r$ is used as the abstract representation of the query-response pair. $\mathbf{v}_q \oplus \mathbf{v}_r$ is further passed through two fully-connected layers. The output of the last fully-connected layer is the estimated probability of the query-response pair being human-generated.
The objective of the discriminator is formalized as follows:
\begin{equation} \label{eq:cross_entropy}
\begin{aligned}
\min_{\phi} & -{\mathbb{E}_{\langle{q, r}\rangle \sim p_{data}} \left[ {\log \left(D_{\phi}(\langle{q,r}\rangle)\right)} \right]} \\
&- {\mathbb{E}_{\langle{q, r}\rangle \sim G_{\theta}} \left[{\log \left(1 - D_{\phi}(\langle{q,r}\rangle)\right)} \right]}
\end{aligned}
\end{equation}
where $p_{data}$ denotes the real-world query-response distribution.
For the response generation task, $D_{\phi}$ is $D_{\phi_{qr}}$ and $G_{\theta}$ is $G_{\theta_{qr}}$, while for the query generation task, $D_{\phi}$ is $D_{\phi_{rq}}$ and $G_{\theta}$ is $G_{\theta_{rq}}$.
\subsection{Dual Generators} \label{sect:model:gen}
Both generators adopt the conventional encoder-decoder Seq2Seq structure, in which GRU is used as the basic unit.
The correlation between the dual tasks~(query generation and response generation) can be represented with the joint probability $P(q,r)$:
\begin{equation} \label{eq:duality_qa_qg}
P(q,r) = P_q(q)P(r|q;\theta_{qr}) = P_r(r)P(q|r;\theta_{rq}) ,
\end{equation}
where $P_q(q)$ and $P_r(r)$  are language models pre-trained on the query corpus and the response corpus. In this paper, we use smoothed bigram language models for both  $P_q(q)$ and $P_r(r)$. $P(r|q; \theta_{qr})$ and $P(q|r; \theta_{rq})$ are the dual generators.
Both $P(r|q; \theta_{qr})$ and $P(q|r; \theta_{rq})$ can be obtained through the markov chain rule:
\begin{equation*}
\begin{cases}
P(r|q; \theta_{qr}) = \prod_{t=1}^{\lvert{r}\rvert}{P(r_t|r_{0:t-1},q;\theta_{qr})} \\
P(q|r; \theta_{rq}) = \prod_{t=1}^{\lvert{q}\rvert}{P(q_t|q_{0:t-1},r;\theta_{rq})}
\end{cases}
\end{equation*}
where $P(r_t|r_{0:t-1},q;\theta_{qr})$ and $P(q_t|q_{0:t-1},r;\theta_{rq})$ are the formulation of the decoders in the Seq2Seq model.
\subsection{Duality Promotes Diversity} \label{sect:model:dual}
To better illustrate why duality increases the diversity of the generated responses, 
we show some query-response pair examples in Figure 2(a).
In Figure~\ref{fig:dual:corpus}, each directional arrow starts from a query while ends at its corresponding response. It can be observed that: (1) Safe response $r_1:$ ``I don't know" connects to many queries, i.e., \{$q_1, q_2, q_3, \cdots$\}. (2) More diverse and specific response $r_2:$ ``The Indian cuisine around the corner is great", nevertheless, exactly corresponds to only one query $q_3:$ ``Where to have dinner?". \footnote{There may exist several other queries that can be replied using ``The Indian cuisine around the corner is great".  But this number is much smaller than those that can be replied using  ``I don't know".  For simplicity, we only show only one query here for the response ``The Indian cuisine around the corner is great". This would not affect the following analysis.  }  


In the training process of $G_{\theta_{rq}}$, the increase of $\log P(q_3 \vert r_2; \theta_{rq})$, denoted by $\Delta \log P(q_3 \vert r_2; \theta_{rq})$, is much bigger than the increase of $\log P(q_3 \vert r_1; \theta_{rq})$, denoted by $\Delta \log P(q_3 \vert r_1; \theta_{rq})$. Formally,
\begin{equation*}
\Delta \log P(q_3 \vert r_2; \theta_{rq}) \gg \Delta \log P(q_3 \vert r_1; \theta_{rq})
\end{equation*}
The reason behind this phenomenon is as follows. The safe response $r_1$ relates with queries $\{ q_1, q_2, q_3, \cdots \}$. When $G_{\theta_{rq}}$ is provided with $\langle{q_1, r_1}\rangle$ or $\langle{q_2, r_1}\rangle$, $G_{\theta_{rq}}$ is optimized to increase the log conditional probability $\log P(q_1 \vert r_1; \theta_{rq})$ or $\log P(q_2 \vert r_1; \theta_{rq})$, it is inevitable that $\log P(q_3 \vert r_1, \theta_{rq})$ will decrease to a certain extent, since these log conditional probabilities share the same parameters $\theta_{rq}$. 
The same principle applies to $\log P(q_2 \vert r_1, \theta_{rq})$ when $G_{\theta_{rq}}$ is provided with $\langle{q_1, r_1}\rangle$ or $\langle{q_3, r_1}\rangle$.   
However, the diverse response $r_2$ is uniquely connected to the query $q_3$, in that case, $G_{\theta_{rq}}$ takes all efforts to increase $\log P(q_3 \vert r_2, \theta_{rq})$. 

With the duality constraint in Eq.~\ref{eq:duality_qa_qg}, we obtain:
\begin{equation} \label{eq:dual:frac}
\frac{P(q|r;\theta_{rq})}{P(r|q;\theta_{qr})}=\frac{P_{q}(q)}{P_r(r)} = k(q,r).
\end{equation}
Since both $P_{q}(q)$ and $P_{r}(r)$ are obtained from the pre-trained language models, both of them are constant for any query-response pair $\langle{q,r}\rangle$.
$k(q,r) = \frac{P_{q}(q)}{P_r(r)}$ is also constant for any $\langle q,r \rangle$.
Take the $\log$ formulation of Eq.~\ref{eq:dual:frac}, we can obtain:
\begin{equation*} \label{eq:dual:log}
\log P(q|r;\theta_{rq}) - \log P(r|q;\theta_{qr})= \log k(q,r).
\end{equation*}
From above equation, we observe that the increase of $\log P(q|r;\theta_{rq})$, denoted as $\Delta \log P(q|r;\theta_{rq})$, and the increase of $\log P(r|q;\theta_{qr})$, denoted by $\Delta \log P(r|q;\theta_{qr})$, is supposed to be equal for any query-response pair $\langle q,r\rangle$, since $\log k(q,r)$ is constant during the training process.
Therefore, 
\begin{equation*}
\Delta \log P(q_3 \vert r_2; \theta_{rq}) \gg \Delta \log P(q_3 \vert r_1; \theta_{rq})
\end{equation*}
in turn makes
\begin{equation*}
\Delta \log P(r_2 \vert q_3 ; \theta_{qr}) \gg \Delta \log P(r_1 \vert q_3; \theta_{qr}).
\end{equation*}
When $G_{\theta_{qr}}$ finishes its training process, we obtain $P(r_2 \vert q_3 ; \theta_{qr}) \gg P(r_1 \vert q_3; \theta_{qr})$. This indicates that it is more likely for $G_{\theta_{qr}}$ to assign higher probability to the diverse response given the query.

We use Figure~\ref{fig:dual:graph} to visually explain this intuition. We suppose that both queries and responses ``possess" their own spatial space. The coordinates of the ellipse and the rectangle represent the locations of the query $q$ and the response $r$ in the spatial space. 
The distance between  $q$ and $r$ represents the probability of transforming between $q$ and $r$, namely $P(q \vert r)$ and $P(r \vert q)$. The shorter the distance, the larger the probability. When $G_{\theta_{qr}}$ and $G_{\theta_{rq}}$ are provided with a query-response pair $\langle{q, r}\rangle$, the training objectives of $G_{\theta_{qr}}$ and $G_{\theta_{rq}}$ are to increase the probability $P(r \vert q)$ and $P(q \vert r)$, i.e., to shorten the distance between $q$ and $r$. 
Since the safe response $r_1$ corresponds to $\{q_1, q_2, q_3, \cdots \}$, the position of this safe response is determined by all involved queries. Because each of these involved queries attempts to ``drag" $r_1$ close to itself, the safe response $r_1$ ``chooses" to keep a distance with each of them to balance the involved queries. However, the diverse response $r_2$ corresponds to exactly one query $q_3$. $r_2$ ``selects" to stay as close to $q_3$ as possible. 
As it can be seen from the figure, the distance between $q_3$ and $r_2$ is much shorter than the distance between $q_3$ and $r_1$, i.e., $P(r_2 \vert q_3)$ is much larger than $P(r_1 \vert q_3)$. In other words, with the duality constraint, $G_{\theta_{qr}}$ tends to generate diverse responses rather than safe responses. 
\section{Training of DAL} \label{sect:training}
\subsubsection*{Duality Constraint for Diversity}
Direct enforcement of the constraint in Eq.~\ref{eq:duality_qa_qg} is intractable.  The duality constraint in Eq.~\ref{eq:duality_qa_qg} can be transformed into a regularization term:
\begin{equation}
\begin{aligned}
\Upsilon &= [\log P_r(r) + \log P(q|r; \theta_{rq}) \\
		&- \log P_q(q) - \log P(r|q;\theta_{qr})]^2
\end{aligned}.
\end{equation}
We minimize $\Upsilon$ to enforce the duality constraint in order to generate more diverse responses.
\subsubsection*{Adversarial Signal for Naturalness} 
The decoding phase in the Seq2Seq model involves sampling discrete words. This discrete sampling makes the optimization of the generator based upon the discriminator's guidance non-differentiable.
To circumvent the non-differentiable obstacle, we optimize each generator through reinforcement learning. The policy gradient is applied to pass the discriminator's adversarial signal to the generator.
The discriminator  $D_{\phi}$ gives a score $J({\theta})$ based on its judgment of how likely the generated $\langle {q,r} \rangle$ is human-generated:
\begin{equation*}
J(\theta) = \mathbb{E}_{{\langle{x, y}\rangle} \in G_{\theta}} [D_{\phi}({\langle{x, y}\rangle})].
\end{equation*}
For response generation, $J(\theta)$ is $J(\theta_{qr})$, $G_{\theta}$ is $G_{\theta_{qr}}$, $D_{\phi}$ is $D_{\phi_{qr}}$, $x$ is the real query and $y$ is the generated response.
Analogously, in query generation, $J(\theta)$ is $J(\theta_{rq})$, $G_{\theta}$ is $G_{\theta_{rq}}$, $D_{\phi}$ is $D_{\phi_{rq}}$, $x$ is the real response and $y$ is the generated query.
$J({\theta})$ is used as the reward for the optimization of $G_{\theta}$.
With the likelihood ration trick~\cite{williams1992simple,sutton2000policy}, the gradient of $J(\theta)$ can be approximated as:
\begin{equation*}
\nabla_{\theta} J(\theta) \simeq [D_{\phi}({\langle{x, y}\rangle}) - b] \cdot \nabla_{\theta} \log (p(y \vert x;\theta)),
\end{equation*}
where $b$ is used to reduce the variance of the estimation while keeping the estimation unbiased, and  $p(y \vert x;\theta)$ is the probability distribution defined by the generator $G_{\theta}$.
\subsubsection*{Combined Gradient}
In DAL, the gradient for updating each generator is the weighted combination of $\nabla_{\theta}J(\theta)$~(for natural responses) and $\nabla_{\theta} \Upsilon$~(for avoidance of safe responses):
\begin{equation} \label{eq:g_gradient}
\begin{cases}
\nabla_{\theta_{qr}} G_{\theta_{qr}} =  \nabla_{\theta_{qr}} \Upsilon -
\lambda_{qr} \cdot \nabla_{\theta_{qr}}J(\theta_{qr})
\\
\nabla_{\theta_{rq}} G_{\theta_{rq}} = \nabla_{\theta_{rq}} \Upsilon -
\lambda_{rq} \cdot \nabla_{\theta_{rq}}J(\theta_{rq})
\end{cases}.
\end{equation}
\subsubsection*{Teacher Forcing} 
When the generator is trained with only the adversarial signals from the discriminator and the duality constraint, the training process of the generator easily collapses. This is because the discriminator sometimes is remarkably better than the corresponding generator in certain training batches. The discriminator can easily discriminate all the generated utterances from real ones. The generator realizes that it generates low-quality samples but cannot figure out the good standard.
To stabilize the training process, 
after each update with the combined gradient $\nabla_{\theta_{qr}} G_{\theta_{qr}}$ or $\nabla_{\theta_{rq}} G_{\theta_{rq}}$, the generators are provided with real query-response pairs and are strengthened with maximum likelihood training, which is also known as Teacher Forcing~\cite{li2017adversarial,lamb2016professor}.
\begin{algorithm}[h]
	\renewcommand{\algorithmicrequire}{\textbf{Input:}}
	\renewcommand{\algorithmicensure}{\textbf{Output:}}
	\caption{Training of DAL.}
	\label{algo:double_discriminators}
	\begin{algorithmic}[1]
			\REQUIRE Two language models $P_q(q)$ and $P_r(r)$ pre-trained on the query corpus and the response corpus.
			\ENSURE $G_{\theta_{qr}}$ and $G_{\theta_{rq}}$
			\STATE Randomly initialize $G_{\theta_{qr}}$,$G_{\theta_{rq}}$, $D_{\phi_{qr}}$, $D_{\phi_{rq}}$.
			\STATE Pre-train $G_{{\theta}_{qr}}$ and $G_{{\theta}_{rq}}$ using maximum likelihood estimation objective.
			\STATE Pre-train $D_{{\phi}_{qr}}$ and $D_{{\phi}_{rq}}$ by Eq.~\ref{eq:cross_entropy}.
			\WHILE {models have not converged}
			\FOR{$i=1,\cdots,d$}
				\STATE Sample $\langle{q, r}\rangle$ from real-world data.
				\STATE Update $D_{{\phi}_{qr}}$ by Eq.~\ref{eq:cross_entropy} with $\langle{q,r}\rangle \sim p_{data}$ and $\langle{q,\hat{r}}\rangle \sim G_{\theta_{qr}}$.
				\STATE Update $D_{{\phi}_{rq}}$ by Eq.~\ref{eq:cross_entropy} with $\langle{q,r}\rangle \sim p_{data}$ and $\langle{\hat{q},{r}}\rangle \sim G_{\theta_{rq}}$.
			\ENDFOR		
			\FOR{$j=1, \cdots, g$}
				\STATE Sample $\langle{q, r}\rangle$ from real-world data.
				\STATE Update $G_{\theta_{qr}}$ by $\nabla_{\theta_{qr}} G_{\theta_{qr}}$ in Eq.~\ref{eq:g_gradient}.
				\STATE Teacher Forcing: update $G_{\theta_{qr}}$with $\langle{q,r}\rangle$
				\STATE Update $G_{\theta_{rq}}$ by $\nabla_{\theta_{rq}} G_{\theta_{rq}}$ in Eq.~\ref{eq:g_gradient}.
				\STATE Teacher Forcing: update $G_{\theta_{rq}}$ with $\langle{q,r}\rangle$
			\ENDFOR
		\ENDWHILE
	\end{algorithmic}
\end{algorithm}

The training procedure of DAL is presented in Algorithm~\ref{algo:double_discriminators}.
Firstly, we use maximum likelihood estimation to pre-train $G_{\theta_{qr}}$ and $G_{\theta_{rq}}$. Analogously, $D_{\phi_{qr}}$ and $D_{\phi_{rq}}$ are also pre-trained according to Eq.~\ref{eq:cross_entropy}.
After the pre-training phase, each generator is optimized by both duality constraint and adversarial signal, followed with the regularization of Teacher Forcing. The corresponding discriminators are simultaneously optimized.
\section{Experiments} \label{sect:experiment}
\subsection{Experimental Settings} \label{sect:experiment:settings}
A Sina Weibo dataset~\cite{zhou2017emotional} is employed to train the models. We treat each query-response pair as a single-turn conversation. Attention mechanism~\cite{luong2015effective} is applied in all the methods to enhance the performance. 
All the methods are implemented based on the open source tools Pytorch\cite{paszke2017automatic} and OpenNMT~\cite{klein2017opennmt}.  Our experiments are conducted on a Tesla K40 cluster. For better replication, we detail the experiment settings, model parameters and preprocessing strategies in the appendix document. Further, we will release our code to the open-source community after the anonymous paper period.

 In order to verify the effectiveness of DAL, we compare the following methods: \\
\noindent$\bullet$ \textit{Seq2Seq:} the standard Seq2Seq model~\cite{sutskever2014sequence}.\\
\noindent$\bullet$ \textit{MMI-anti:} the mutual information method~\cite{li2016diversity}, which uses an anti-language model in inference.  \\
\noindent$\bullet$ \textit{MMI-bidi:} the mutual information method~\cite{li2016diversity}, which first generates a N-best response set with $p(r \vert q)$ and then reranks this response set with $p(q \vert r)$ in inference. \\
\noindent$\bullet$ \textit{Adver-REIN:} the adversarial method adopting REINFORCE algorithm~\cite{li2017adversarial}.\\
\noindent$\bullet$ \textit{GAN-AEL:} the adversarial method with an approximate embedding layer to solve the non-differentiable problem~\cite{xu2017neural}.\\
\noindent$\bullet$ \textit{DAL-Dual~(ours):} DAL trained only with maximum likelihood~(Teacher Forcing) and duality constraint~($\nabla_{\theta_{qr}} \Upsilon$ or $\nabla_{\theta_{rq}} \Upsilon)$. \\
\noindent$\bullet$ \textit{DAL-DuAd~(ours):} \textit{DAL-Dual} with adversarial learning~(Algorithm~\ref{algo:double_discriminators}).
\noindent 

Both \textit{DAL-Dual} and \textit{DAL-DuAd} are methods proposed by us: the former incorporates the dual signal only, while the later combines the dual signal and the adversarial signal.  
\subsection{Experimental Results} \label{sect:experiment:results}
We firstly evaluate DAL on the task of generating of diverse responses. Then we resort to human annotators to evaluate the overall quality of the generated responses. Finally, we present several cases generated by all the involved method. 
\subsubsection*{Response Diversity Measured by Distinct}
DISTINCT is a well-recognized metric to evaluate the diversity of the generated responses~\cite{li2016diversity,xing2017topic}. In our experiment, we employ DISTINCT-1 and DISTINCT-2, which calculate distinct unigrams and bigrams in the generated responses respectively.
Table~\ref{tab:weibo_Q2R_autoeval} presents the results of the five methods. 
\begin{table}[ht!]
\centering
\resizebox {0.85\columnwidth} {!} 
{
	\begin{tabular}{|c|c|c|}
		\hline
		{Method} & DISTINCT-1 & DISTINCT-2 \\
		\hline
		Seq2Seq & 0.031 & 0.137 \\
		MMI-anti & 0.033 & 0.141 \\
		MMI-bidi & 0.034 & 0.143  \\
		Adver-REIN & 0.036 & 0.145  \\
		GAN-AEL & 0.038 & 0.149 \\
		\textbf{DAL-Dual~(\textit{ours})} & \textbf{0.052} & \textbf{0.209}  \\		
		\textbf{DAL-DuAd~(\textit{ours})} & \textbf{0.049} & \textbf{0.201} \\
		\hline	

		\hline		
	\end{tabular}	
}
\caption{\label{tab:weibo_Q2R_autoeval} Results of diversity evaluation.}
\end{table}

From Table~\ref{tab:weibo_Q2R_autoeval}, we have the following observations:
(1) Both \textit{MMI-anti} and \textit{MMI-bidi} slightly improve the performance as compared with \textit{Seq2Seq}. 
\textit{MMI-bidi} heavily relies on the diversity of the N-best response set generated by $p(r \vert q)$.
When $N$ is not large enough to include some infrequently-occurring responses into the optional set, this set may lack diversity, and thus the ultimate response obtained with the reranking strategy also lacks diversity.
However, when $N$ is large, some responses having low coherence with the given query will be included in the optional set, and such responses may be selected as the final response, which hurts the performance of \textit{MMI-bidi}. Therefore, the selection of $N$ is an arduous task. 
 \textit{MMI-anti} also heavily relies on the anti-language model to obtain diverse responses. 
(2) Compared with \textit{Seq2Seq}, our \textit{DAL-Dual} improves diversity by 67.7\% measured by DISTINCT-1 and 52.6\% measured by DISTINCT-2, which reveals the effectiveness of the dual approach in improving diversity.
(3) As expected, compared with \textit{Adver-Rein} and \textit{GAN-AEL}, our \textit{DAL-DuAd} further improves the diversity of the generated responses. This observation proves our assumption that, with the guidance of discriminators $D_{\phi_{qr}}$ and $D_{\phi_{rq}}$, the generator $G_{\theta_{rq}}$ is able to influence the generator $G_{\theta_{qr}}$ to produce more diverse responses.

We do notice that \textit{DAL-Dual} achieves slightly better performance than \textit{DAL-DuAd} on diversity. The reason is that sometimes adversarial methods tend to generate some short but quality responses such as ``Let's go!'' for given queries such as ``We can have dinner together tonight. '' or ``There is an exhibition at the National Museum.''. However, this short but natural response would harm diversity metrics. However, this does not violate the assumption that adversarial signal makes generated responses more natural. To demonstrate further boost brought by adversarial signal, we conduct two pair-wise experiments in appendix file.
\subsubsection*{Response Quality Evaluated by Human}
\begin{figure*}[h!]
\centering
\includegraphics[width=0.99\textwidth]{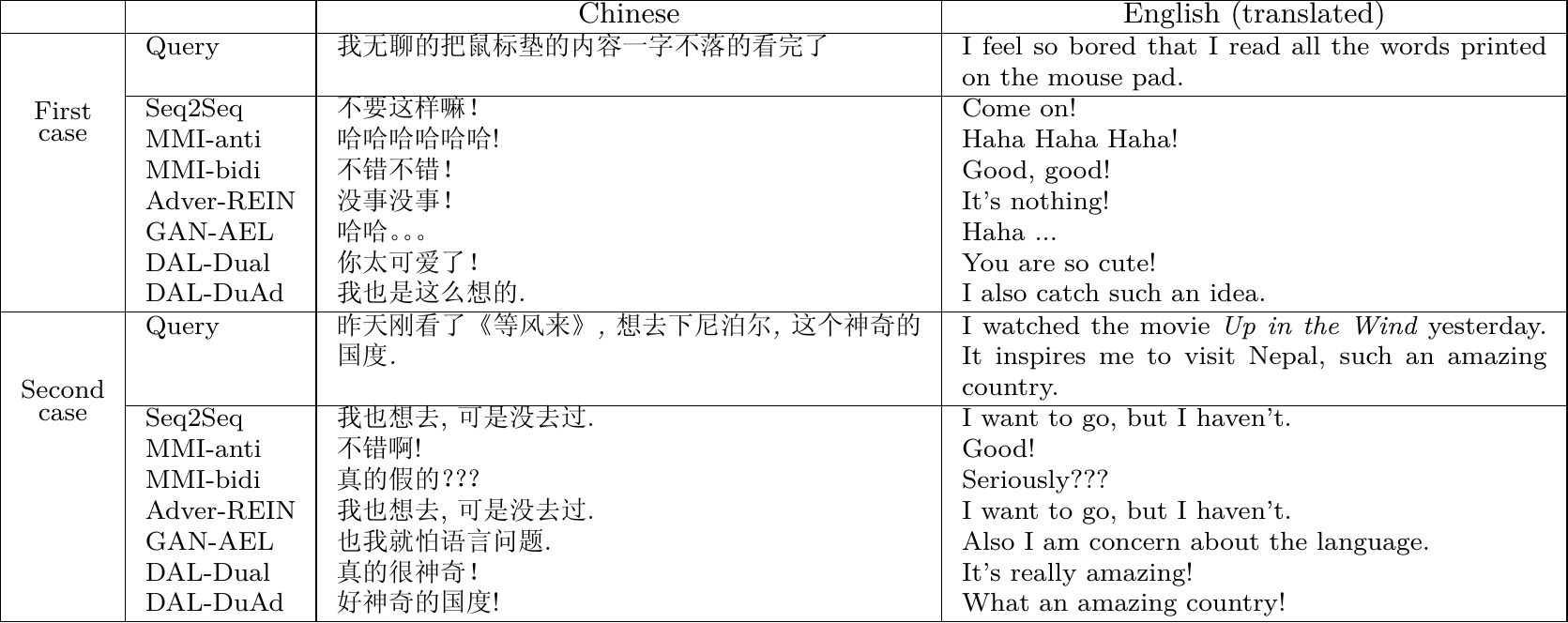}
\caption{\label{tab:case} Case study.}
\end{figure*}
Since the word overlap-based metrics such as BLEU~\cite{papineni2002bleu} and embedding-based metrics are inappropriate for response quality evaluation due to their low correlation with human judgment~\cite{liu2016not,mou2016sequence}, we resort to human annotators to evaluate the overall quality of the generated responses. We employ annotators to evaluate the quality of 200 responses generated from each of the aforementioned methods.
Three annotators are required to score the overall quality of the generated responses. \textbf{2}: the response is natural, relevant and informative. \textbf{1}: the response is appropriate for the given query but may not be very informative. \textbf{0}: the response is completely irrelevant, incoherent or contains syntactic errors. The final score for each response is the average of the scores from all the annotators. The human evaluation results are listed in Table~\ref{tab:weibo_Q2R_human_pointwise}.  
\begin{table}[ht!]
\centering
\resizebox {0.8\columnwidth} {!} {
	{
	\begin{tabular}{|c|cc|c|}
		\hline
		{Method} & Human rating & Kappa\\
		\hline
		Seq2Seq & 0.470 & 0.56\\
		MMI-anti & 0.568 & 0.46 \\
		MMI-bidi & 0.523  & 0.60 \\
		Adver-REIN & 0.767 & 0.49\\
		GAN-AEL & 0.758 & 0.52 \\
		DAL-Dual~(\textit{ours}) & 0.730 & 0.47\\
		\textbf{DAL-DuAd~(\textit{ours})} & \textbf{0.778}  & 0.50 \\
		\hline
	\end{tabular}		
	}
}
\caption{\label{tab:weibo_Q2R_human_pointwise} Results of human elevation: response quality.}
\end{table}

The agreement among annotators is calculated with Fleiss' kappa~\cite{fleiss1971measuring}. The agreement ratio is in a range from 0.4 to 0.6, showing moderate agreement.
Based on the results, we have the following observations:
(1) \textit{DAL-DuAd} achieves the highest quality score, indicating that our \textit{DAL-DuAd} has the ability to produce coherent and informative responses.
(2) \textit{Adver-REIN} and \textit{GAN-AEL} also obtain  fairly good pointwise scores. This is because the adversarial learning mechanism effectively guides the generated responses to be close to the human-generated responses.
(3) Compared with \textit{Seq2Seq}, \textit{MMI-anti} and \textit{MMI-bidi}, our \textit{DAL-Dual} obtains relatively satisfactory performance on overall quality. It shows that the dual signal can also improve the overall quality. 
\subsubsection*{Pairwise Experiment}
We conduct the pairwise evaluation on \{\textit{Seq2Seq}, \textit{DAL-Dual}\} and \{\textit{DAL-Dual}, \textit{DAL-DuAd}\}. The former aims to evaluate the dual signal while the latter targets on assessing the adversarial signal. 200 queries are used to evaluate the methods. The comparison results are shown in Table~\ref{tab:weibo_Q2R_human_pairwise}. 
The annotator agreement ratio is also in a range from 0.4 to 0.6, which is interpreted as moderate agreement~\cite{fleiss1971measuring}.
The comparison between \{\textit{Seq2Seq}, \textit{DAL-Dual}\} shows that \textit{DAL-Dual} outperforms \textit{Seq2Seq}, which demonstrates the effectiveness of the dual signal in improving the overall quality.
Furthermore, the comparison between \textit{\{DAL-Dual, DAL-DuAd\}} proves that the guidance from the discriminator can further improve the overall quality. The pairwise results verify that the dual signal and the adversarial signal in  DAL collaborate together to enhance the overall quality of the generated responses.
\begin{table}[H]
\centering
\resizebox {\tablewidth\columnwidth} {!} {
	{
	\begin{tabular}{|c|ccc|c|}
		\hline
		{Method} & Wins  & Ties & Losses & Kappa \\
		\hline
		Seq2Seq & 24.25\% & 45.00\% & 30.75\% & \multirow{2}{*}{0.47}\\
		DAL-Dual & 30.75\% & 45.00\% & 24.25\% & \\
		\cline{1-5}
		DAL-Dual & 23.50\% & 49.00\% & 27.50\% & \multirow{2}{*}{0.49}\\
		DAL-DuAd & 27.50\% & 49.00\%  & 23.50\% & \\
		\hline
	\end{tabular}		
	}
}
\caption{\label{tab:weibo_Q2R_human_pairwise} Results of human elevation: \textit{pairwise}.}
\end{table}
\subsubsection*{Case Study} 
We present several cases in Figure~\ref{tab:case}. 
For the first case involving \textit{the content on the mouse pad}, most of the baselines generate generic responses  such as``Come on!", ``Haha!" or ``It's nothing!". 
On the contrary, our \textit{DAL-Dual} and \textit{DAL-DuAd} method produce much more diverse and informative responses, such as ``You are so cute!" and ``I also catch such an idea.". These two entertaining responses are also topically coherent and logically consistent with the given query. 
In the second cases, our methods are also capable of capturing the topic \textit{amazing country} shown in the query, and well generate the diverse and coherent responses following the topic of the query, such as ``What an amazing country!" or ``It is really amazing!". In contrast, the baselines still tend to provide safe responses lacking diversity to different queries.
\subsection{Comparison of Efficiency} \label{sect:experiment:efficiency}
\begin{figure}[ht!]
\centering
\includegraphics[width=0.9\columnwidth]{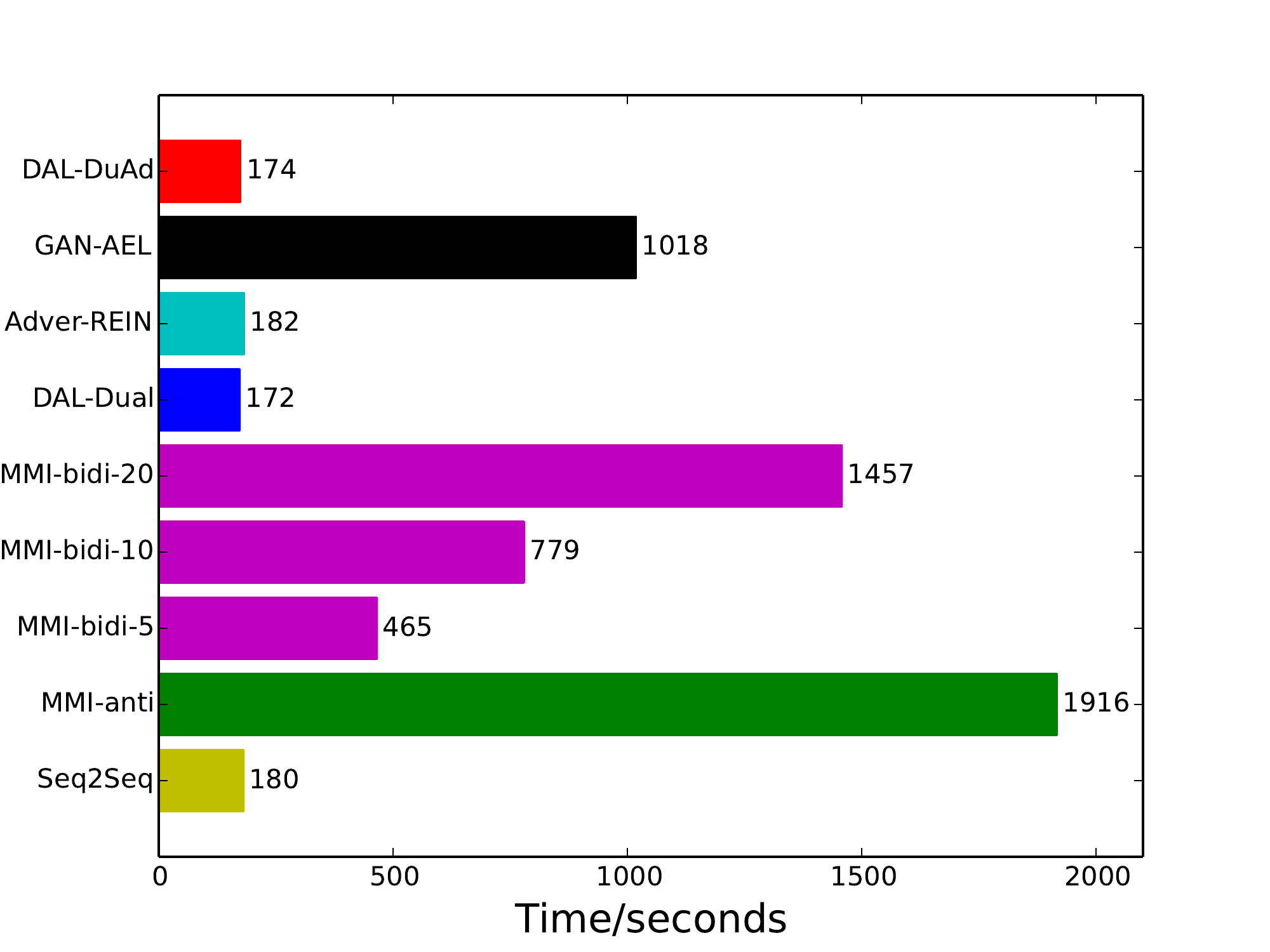}
\caption{\label{fig:efficiency} Time consumed by different methods.}
\end{figure}

Efficiency is a crucial factor for real-life applications such as online chatbots. We conduct an experiment to evaluate the efficiency of all the methods under study. 
The efficiency experiment is conducted ten times on one Tesla K40m GPU whose memory is 11471M. 
The average time consumed by each method to generate the responses for 1000 queries is reported in Figure~\ref{fig:efficiency}. \textit{MMI-bidi-5}, \textit{MMI-bidi-10} and \textit{MMI-bidi-20} denote the \textit{MMI-bidi} method with the N-best size of 5, 10 and 20 respectively. 
We can see that \textit{MMI-anti} and \textit{GAN-AEL} are the most time-consuming in all the baselines.  
Besides, we note that \textit{MMI-bidi} method with the reranking strategy, even with a relatively small N-best size of 5, consumes much longer time than our methods,  which severely limits \textit{MMI-bidi}'s application in practice.
However, \textit{Seq2Seq}, \textit{Adver-REIN}, \textit{DAL-Dual} and \textit{DAL-DuAd} have very similar efficiency performance. 
Compared with \textit{Seq2Seq} and \textit{Adver-REIN}, \textit{DAL-Dual} and \textit{DAL-DuAd} achieve much better performance on diversity and overall quality.
Therefore, DAL is more suitable for real-life applications.  
\section{Conclusion} \label{sect:conclusion}
We  propose a novel  framework named DAL
to alleviate two prominent problems~(safe responses and unnatural responses) plaguing dialogue generation.
The dual learning proposed in this paper is the first effort to utilize the reverse dependency between queries and responses to reduce the probability of safe response generation and improve the diversity of the generated responses. Adversarial learning makes the generated responses as natural to human-generated ones as possible. DAL seamlessly integrates dual learning and adversarial learning, which are complementary to each other. 
Experimental results show that DAL achieves better performance than the state-of-the-art methods in terms of diversity, overall quality and efficiency. 
\bibliography{naaclhlt2019}
\bibliographystyle{acl_natbib}

\end{document}